\documentclass{article}
\usepackage{spconf,amsmath,graphicx, color}
\usepackage{multirow}
\usepackage{amssymb}
\usepackage{soul}
\usepackage{makecell}
\usepackage{color}

\title{Lightweight network towards real-time image denoising on mobile devices}
\name{Zhuoqun Liu$^{\dagger \star}$ Meiguang Jin$^{\dagger}$ Ying Chen$^{\dagger}$ Huaida Liu$^{\dagger}$ Canqian Yang$^{\dagger}$ Hongkai Xiong$^{\star}$}

\address{$^{\dagger}$ Alibaba Group \\
$^{\star}$ Shanghai Jiao Tong University}
\begin{document}

\maketitle
%
\begin{abstract}

Deep convolutional neural networks have achieved great progress in image denoising tasks. However, their complicated architectures and heavy computational cost hinder their deployments on mobile devices. Some recent efforts in designing lightweight denoising networks focus on reducing either FLOPs (floating-point operations) or the number of parameters. However, these metrics are not directly correlated with the on-device latency. In this paper, we identify the real bottlenecks that affect the CNN-based models' run-time performance on mobile devices: memory access cost and NPU-incompatible operations, and build the model based on these. To further improve the denoising performance, the mobile-friendly attention module \textit{MFA} and the model reparameterization module \textit{RepConv} are proposed, which enjoy both low latency and excellent denoising performance. To this end, we propose a mobile-friendly denoising network, namely \textit{MFDNet}. The experiments show that \textit{MFDNet} achieves state-of-the-art performance on real-world denoising benchmarks SIDD and DND under real-time latency on mobile devices. The code and pre-trained models will be released.
\end{abstract}
\section{Introduction}\vspace{-0.2cm}
\label{sec:intro}

With the rapid development of deep learning techniques, the performance of image denoising is improved significantly in recent years \cite{chen2022simple, zamir2022restormer, chen2021hinet, wang2022uformer}. However, deploying a state-of-the-art (SOTA) denoising model on resource constrained devices, such as mobile devices, remains challenging. 
On the one hand, although NPUs specifically optimized for deep neural networks are ubiquitous on mobile devices, most SOTA network architectures do not consider NPUs' compatibility. Hence they cannot fully utilize the powerful NPUs. 
On the other hand, the requirement of high resolution processing (720p/1080p or even higher) for real applications exponentially increases the computational and memory access cost, which are key efficiency bottlenecks on mobile devices.
There have been some attempts to design lightweight models to promote mobile deployments \cite{xu2021efficient, zamir2022restormer, chen2022simple}. These lightweight models reduce either the FLOPs or the number of parameters of the model. However, recent works show that reducing the FLOPs or the number of parameters does not necessarily lead to a low latency on mobile devices \cite{vasu2022improved, zhang2021edge}. 
For example, skip connections and multi-branch structures are commonly used design choices for low-level vision tasks \cite{MPRnet, wang2022uformer, chen2021hinet, zamir2022restormer}. However, these operations can incur a high memory access cost, hindering fast inference on mobile devices. In addition to the model architectures, whether the operations are well optimized by NPUs is also essential when improving runtime performance \cite{zhang2021edge}. Benefiting from the powerful parallel computing capability and specialized optimization for common operations, NPUs show great advantages over other processors when processing neural networks. However, operations well optimized by the NPUs are quite limited. Architectures containing NPU-incompatible operations will be partially processed by CPUs or GPUs. This introduces the additional data transfer cost between processors and increases the synchronization cost, leading to a severe overhead. For example, the ESA module \cite{ESA}, which is adopted by the winner in the runtime track of the NTIRE 2022 efficient super-resolution challenge \cite{ESA-ntire}, contains NPU-incompatible operations like max-pooling with a kernel size of 7 and a stride of 3 and bilinear interpolation with a scale larger than 2, which will be processed by GPUs. In this case, the feature maps have to be moved from NPUs to GPUs until the process is finished, which significantly slows down the inference speed of the model, as shown in Table \ref{attention performance}.

In this paper, by conducting extensive experiments on an iPhone 11, we identify the preference of the Apple Neural Engine (ANE), which is a typical type of NPUs, for different network architectures and operations. Based on this, a mobile-friendly denoising network is proposed to significantly improve the model's runtime performance on mobile devices while achieving enhanced denoising performance than other lightweight denoising networks on the real-world denoising benchmarks SIDD \cite{sidd} and DND \cite{dnd}. The experiments show that our model only takes around 20ms when processing a 720p image on an iPhone 11, which offers the possibility to process 720p images or videos in real-time on mobile devices.

\section{Network Architecture}\vspace{-0.15cm}


In this section, we build a mobile-friendly denoising network from scratch. To ensure that our model can have efficient runtime performance on mobile devices, we only use operations compatible with the ANE. The results of models with different capacities are shown in the experiment section. \vspace{-0.4cm}

\subsection{Baseline Model} \vspace{-0.1cm}


\begin{figure}[t]
\centering
\includegraphics[width=\columnwidth]{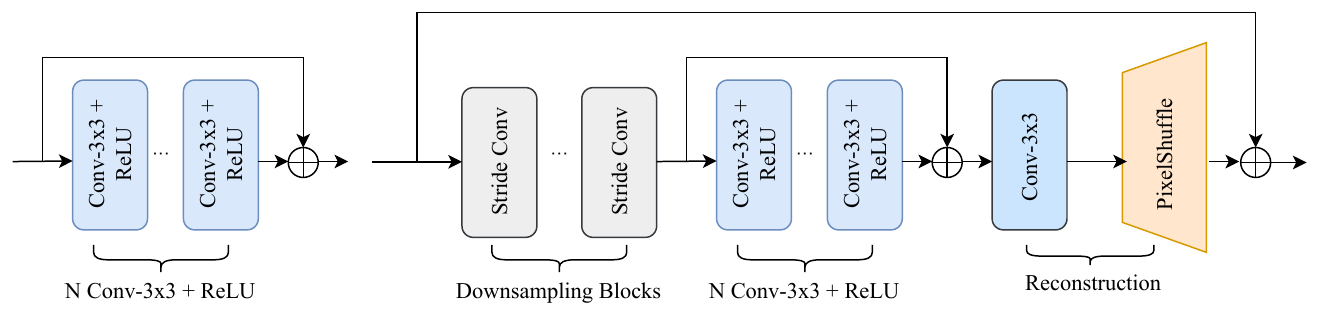}
\caption{The architecture of the baseline model.}
\label{plain_model_with_downsampling}
\end{figure}

\begin{table}[t]\scriptsize
\caption{Quantitative performance of different downsampling factors.}\vspace{0.2cm}
\centering
\label{performance with different scale}
\setlength{\tabcolsep}{1.2mm}{
\begin{tabular}{|c|c|c|c|c|c|c|}
\hline
Model                     & Setting & Factor & MACs/G & Memory/M & PSNR/dB & Latency/ms \\ \hline
\multirow{3}{*}{Baseline} & C16\_N8   & 1      & $12.61$  & $2271$     & $36.74$    & $19.87$       \\ \cline{2-7} 
                          & C32\_N12  & $\downarrow$ 2      & $24.66$  & $1448$     & $\bf{38.52}$    & $18.96$       \\ \cline{2-7} 
                          & C48\_N16  & $\downarrow$ 4      & $20.30$  & $822$      & $\bf{38.52}$    & $\bf{16.87}$       \\ \hline
\end{tabular}}
\end{table}

As mentioned before, to ensure real-time on-device runtime performance, the memory access cost has to be strictly controlled. Therefore, we start from a DnCNN-like \cite{zhang2017beyond} plain topology, as shown in Fig.\ref{plain_model_with_downsampling} left, which contains only the well-optimized $3 \times 3$ convolutions and ReLU activation functions. To reduce the memory access cost, only one residual connection is adopted. We start from this architecture because the DnCNN-like architectures have demonstrated their effectiveness in the image denoising task \cite{zhang2017beyond}. And we remove batch normalization to reduce potential artifacts \cite{wang2018esrgan, lim2017enhanced}.

Based on this plain topology, we adopt multiple $3\times3$ convolutions with stride 2 to downsample the input, followed by $N$ Conv-3$\times$3+ReLU blocks with a local skip connection. At the end of the network is a PixelShuffle \cite{shi2016real} for reconstruction. The modified model is shown in Fig. \ref{plain_model_with_downsampling} right.

Downsampling in the beginning brings two benefits. First, most of the operations in the model are executed at a low resolution, which significantly reduces the total memory access cost, as shown in Table \ref{performance with different scale}. Second, the overall increased receptive field of the network enables the network to capture more contextual information and improves denoising performance. 

We compare the denoising performance of the models with different downsampling factors on the SIDD validation dataset and the runtime performance on an iPhone 11. Both the runtime performance and the total memory access cost (read and write) are evaluated with a 720p input. The results are summarized in Table \ref{performance with different scale}, where $C$ and $N$ represent the number of channels and the number of Conv-3$\times$3+ReLU blocks, respectively. 
Results show that the pre-downsampling enables a wider and deeper model under a similar latency by significantly reducing the memory access cost, thus leading to an enhanced denoising performance. 
So we choose the 4$\times$ downsampling model as our final baseline model.\vspace{-0.4cm}

\subsection{Attention}\vspace{-0.1cm}

\begin{figure}[t]
    \centering
    \includegraphics[scale=0.5]{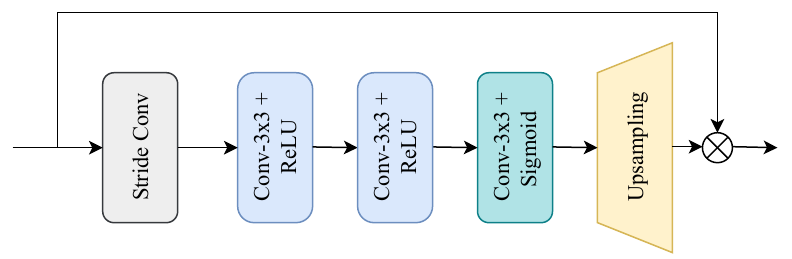}
    \caption{The architecture of MFA.}
    \label{attention}
\end{figure}



Attention mechanisms have been extensively studied in low-level vision tasks \cite{ESA, hfab, chen2022simple, zamir2022restormer, wang2022uformer}. In previous works, attention modules often adopt complex topology or use operations that are not optimized by ANE \cite{ESA}, which affects the runtime performance severely, as shown in Table \ref{attention performance}. In this paper, we propose a simple yet effective mobile-friendly attention module, MFA. Its architecture is shown in Fig.\ref{attention}. Our proposed architecture has three advantages. First, in order to reduce the memory access cost and boost the runtime performance of the module, we discard the complex topology and retain only the necessary residual connection for multiplying the learned attention maps with the input feature maps as a spatial attention mechanism. Second, we further downsample the feature maps to reduce the latency and enlarge the receptive field of the attention module. Finally, we only use operations well-optimized by ANE, like the common $3\times3$ convolution, the ReLU activation function and bilinear interpolation with a scale of 2 to further boost the efficiency. 

 \begin{table}[t]\scriptsize
\caption{Performance of different attention mechanisms.}\vspace{0.2cm}
\centering
\label{attention performance}
\setlength{\tabcolsep}{1.8mm}{
\begin{tabular}{|c|c|c|c|c|c|}
\hline
Model                     & Attention & MACs/G & Memory/M & PSNR/dB & Latency/ms \\ \hline
\multirow{4}{*}{Baseline} & ESA \cite{ESA}      & $12.78$  & $562$      & $38.59$    & $43.23^*$       \\ \cline{2-6} 
                          & SCA \cite{chen2022simple}      & $12.52$  & $400$      & $38.41$    & $\bf{15.20}$       \\ \cline{2-6} 
                          & HFAB \cite{hfab}     & $14.60$  & $574$      & $38.51 $   & $18.94$       \\ \cline{2-6} 
                          & MFA       & $12.85$  & $448$      & $\bf{38.60}$    & $15.61$       \\ \hline
\end{tabular}}
\end{table}

In practice, we propose a mobile-friendly denoising block, MFDB, which contains $K$ Conv-3$\times$3+ReLU blocks followed by one MFA. The width of the MFA is set to $\frac{1}{4}$ of the width of the model. In Table \ref{attention performance}, we compare MFA with the attention modules commonly used in low-level vision tasks, including ESA \cite{ESA}, SCA \cite{chen2022simple}, and HFAB \cite{hfab}. It can be seen from the table that MFA achieves the best denoising performance. Meanwhile, MFA is also close to SCA in terms of latency, with a difference of only 0.4ms. Note that although ESA has a comparable memory access cost compared to other methods, its on-device latency is still significantly higher due to incompatible operations mentioned in section 1.\vspace{-0.3cm}

\subsection{Activation}\vspace{-0.1cm}
Although Rectified Linear Unit (ReLU) has been extensively used in low-level vision tasks, many SOTA methods tend to replace ReLU with other activation functions, such as GELU, LReLU, PReLU \cite{chen2022simple, hfab} for better performance. In order to test the compatibility of the ANE for different activation functions and the potential performance gains, we replace ReLU with several different activation functions, as shown in Table \ref{activation}. The results in the table show that these activation functions are all well-optimized by the ANE. Replacing ReLU with LReLU results in a performance gain of 0.13dB on the validation dataset of SIDD, while the inference time on an iPhone 11 remains nearly unchanged. We, therefore, replace all the ReLU in our baseline model with LReLU.\vspace{-0.2cm}

\begin{table}[t]\scriptsize
\caption{Quantitative performance of different activation functions.} \vspace{0.2cm}
\centering
\label{activation}
\setlength{\tabcolsep}{4.5mm}{
\begin{tabular}{|c|c|c|c|}
\hline
Model                                                                          & Activation & PSNR/dB & Latency/ms \\ \hline
\multirow{4}{*}{\begin{tabular}[c]{@{}c@{}}Baseline+\\ MFA\end{tabular}} & ReLU       & $38.60$    & $15.61$       \\ \cline{2-4} 
                                                                               & GELU       & $38.71$    & $16.11$       \\ \cline{2-4} 
                                                                               & PReLU      & $38.71$    & $15.84$       \\ \cline{2-4} 
                                                                               & LReLU      & $38.73$    & $15.74$       \\ \hline
\end{tabular}}
\end{table}\vspace{-0.2cm}

\subsection{Lightweight Feature Extraction}\vspace{-0.1cm}

To further improve the feature extraction and representation capabilities of the downsampling module, we replace the stride-2 convolution with the Haar transform, which can be efficiently implemented in a convolution form with a kernel size of 2. Compared to a stride-2 convolution, Haar transform is invertible, which ensures that the frequency information of the input can be effectively captured in a lossless manner. 
This is very helpful in preserving 
textures
\cite{liu2018multi}. In addition, Haar transform is a more compact feature representation, meaning that fewer hidden channels between the downsampling blocks can lead to a better denoising performance, which also reduces the latency. Table \ref{ablation} shows that using the Haar transform to extract features significantly improves the model's denoising and on-device runtime performance. It brings 0.22dB performance gain on SIDD and 1.32ms latency reduction on an iPhone 11.\vspace{-0.4cm}

\begin{figure}[b]
    \centering
    \includegraphics[scale=0.56]{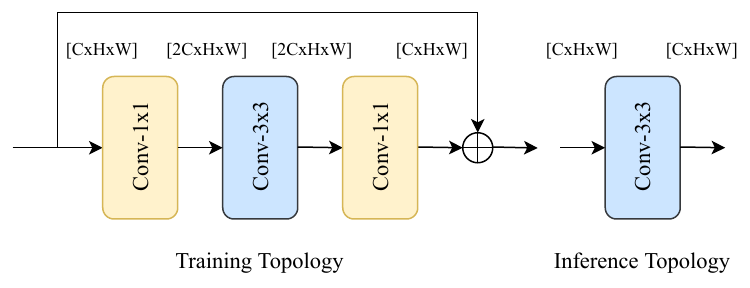}
    \caption{The structure of RepConv.}
    \label{repconv}
\end{figure}

\begin{figure}[t]
    \centering
    \includegraphics[width=\columnwidth]{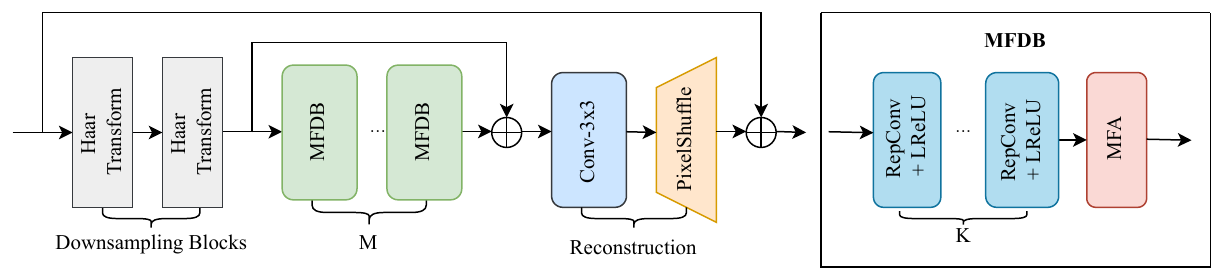}
    \caption{The overall architecture of MFDNet.}
    \label{overall}
\end{figure}

\subsection{Reparameterization}\vspace{-0.1cm}


The idea of reparameterization was first proposed by RepVGG \cite{ding2021repvgg}. The core idea behind reparameterization is to parameterize a plain topology with the parameters transformed from a more complex topology (e.g., multi-branch topology). Model reparameterization is very beneficial in the design of mobile-friendly models. 
Complex topologies, such as multi-branch structures, can significantly increase the memory access cost and slow down the inference. However, the plain topology lags in the feature extraction capability, resulting in compromised model performance. Model reparameterization can be used to address this issue. In the training phase, the model takes advantage of the complex topology to enrich the feature representation and bring performance gains. In the testing phase, the model reparameterization method is used to simplify the topology and improve the inference speed of the model without a performance drop.

In this paper, we use the expand-and-squeeze topology for training, named RepConv, since the wider features result in better feature representation. As shown in Fig.\ref{repconv}, the topology consists of two $1\times 1$ convolutions, a $3\times 3$ convolution, and a skip connection in the training phase. In the testing phase, we merge three convolutions and a skip connection into a single $3 \times 3$ convolution by reparameterization, thus eliminating the cascaded and multi-branch structures. In Table \ref{rep}, we compare the RepConv with other reparameterization methods commonly used in low-level vision tasks, including ECB \cite{zhang2021edge} and RRRB \cite{hfab}. Results show that RepConv achieves the best denoising performance with an improvement of 0.04dB on SIDD.\vspace{-0.3cm}
\begin{table}[t]\scriptsize
\caption{Comparisons between different reparameterization methods.} \vspace{0.2cm}
\centering
\label{rep}
\setlength{\tabcolsep}{4.5mm}{
\begin{tabular}{|c|c|c|}
\hline
Model                                                                                    & Reparameterization Method & PSNR/dB \\ \hline
\multirow{4}{*}{\begin{tabular}[c]{@{}c@{}}MFDNet w/o\\ reparameterization\end{tabular}} &           /               & 38.95    \\ \cline{2-3} 
                                                                                         & ECB \cite{zhang2021edge}                       & 38.97    \\ \cline{2-3} 
                                                                                         & RRRB \cite{hfab}                      & 38.95    \\ \cline{2-3} 
                                                                                         & RepConv                   & 38.99    \\ \hline
\end{tabular}}
\end{table}
\subsection{Summary}


At this point, we build our mobile-friendly denoising network step by step from the baseline model. The architecture of MFDNet is shown in Fig.\ref{overall}. For MFDNet, we set both the number of MFDBs ($M$) and RepConv+LReLU blocks in each MFDB ($K$) to 3.

\section{Experiment}


In this section, we first analyze the role of the different model design choices mentioned in the previous sections in terms of both denoising and runtime performance. We then apply our model to the real-world denoising benchmarks SIDD and DND. To test the models' on-device runtime performance, we execute them 300 times on an iPhone 11 to get the average elapsed time. Note that in all experiments, latency with the $*$ notation indicates that the model contains ANE incompatible operations and is processed by CPUs/GPUs. The computational complexity, latency and total memory access cost in all experiments are evaluated with a 720p input.\vspace{-0.35cm}

\begin{table}[t]\scriptsize
\caption{Ablation study of different components in MFDNet.}\vspace{0.2cm}
\centering
\label{ablation}
\setlength{\tabcolsep}{0.43mm}{
\begin{tabular}{|c|c|c|c|c|c|c|}
\hline
                             & MFA                 & \begin{tabular}[c]{@{}c@{}}ReLU \\  $\rightarrow$ LReLU\end{tabular} & \begin{tabular}[c]{@{}c@{}}Stride-2 Conv \\ $\rightarrow$ Haar Transform\end{tabular} & RepConv        & PSNR/dB &Latency/ms\\ \hline
\multirow{5}{*}{Baseline} &                           &                                                        &                                                                   &                           &$38.54$          & $12.82$                                                                  \\ \cline{2-7} 
                             & \(\checkmark\) &                                               &                                                                   &                           & $38.60$    & $15.61$                                                            \\ \cline{2-7} 
                             & $\checkmark$ & $\checkmark$                              &                                                                   &                           & $38.73$    &$15.74$                                                                  \\ \cline{2-7} 
                             & \checkmark & \checkmark                              & \checkmark                                         &                           & $38.95$    & $14.42$                                                            \\ \cline{2-7} 
                             & \checkmark & \checkmark                              & \checkmark                                         & \checkmark & $38.99$    & $14.42$                                                            \\ \hline
\end{tabular}}
\end{table}

\subsection{Ablation Study}\vspace{-0.1cm}



We conduct our ablation study on the validation dataset of SIDD and measure the models' latency on an iPhone 11. We train our model using the Adam optimizer with the learning rate initialized with 4e-4 and halved every 100k iterations. We train the model for a total of 1M iterations with a batch size of 32 and use cropped patches of 256$\times$256 from SIDD-Medium as the training dataset. 

We start from the baseline model with a downsampling factor of 4, as mentioned in section 3.1, and build the MFDNet step by step. We set the depth of the baseline model ($N$) to 9 and the width to 48. Table \ref{ablation} shows the effectiveness of different components. \vspace{-0.35cm}

\subsection{Application}

We evaluate our model on two real-world denoising benchmarks, SIDD and DND. All training settings are identical to those used in the ablation experiments. 
To see the generalization ability of our model on different NPUs, we also test the on-device latency on iPhone 14 Pro.
The results are summarized in Table \ref{denoising and runtime performance}. For MFDNet, we also design two models with different scales: MFDNet-S and MFDNet-L. For MFDNet-S, we reduce $M$ and $K$ to 1. For MFDNet-L, we keep $K=3$ unchanged and increase $M$ to 6. We compare our model with models proposed in \cite{zhang2017beyond, zhuo2019ridnet, chen2022simple, chen2021hinet, wang2020practical}. Note that as these models are not designed specifically for mobile devices, we prune them in terms of depth and width to ensure that they can run on mobile devices. For DnCNN \cite{zhang2017beyond}, we set both the width and depth to 12, and for DnCNN-S, we set the width to 16 and the depth to 3. For RIDNet \cite{zhuo2019ridnet}, we trim the number of channels to 8 and set the channel reduction to 2. For NAFNet \cite{chen2022simple}, the number of channels is also trimmed to 8 and the number of blocks is reduced to 7. For HINet \cite{chen2021hinet}, we adjust the number of channels and the depth of the model to 8 and 2, respectively. 


We can see from Table \ref{denoising and runtime performance} that, MFDNet and MFDNet-L achieve the best denoising performance with the lowest latency compared to models with comparable computational complexity. For models with computational complexity under 5GMACs, MFDNet-S achieves the best runtime performance with the denoising performance far exceeds that of DnCNN. Note that the proposed MFDNet is fully compatible with both iPhone 11 and iPhone 14 Pro, which indicates its excellent generalization ability. In contrast, NAFNet and HINet are only compatible with iPhone 14 Pro. The layer normalization operation in NAFNet and the instance normalization operation in HINet are incompatible with iPhone 11, which severely affect the inference speed of these models. By removing the incompatible operations from NAFNet and HINet, the latency on iPhone 11 will reduce to 71.3ms and 47.3ms, respectively.\vspace{-0.6cm}
\begin{figure}[t]
    \centering
    \includegraphics[width=\columnwidth]{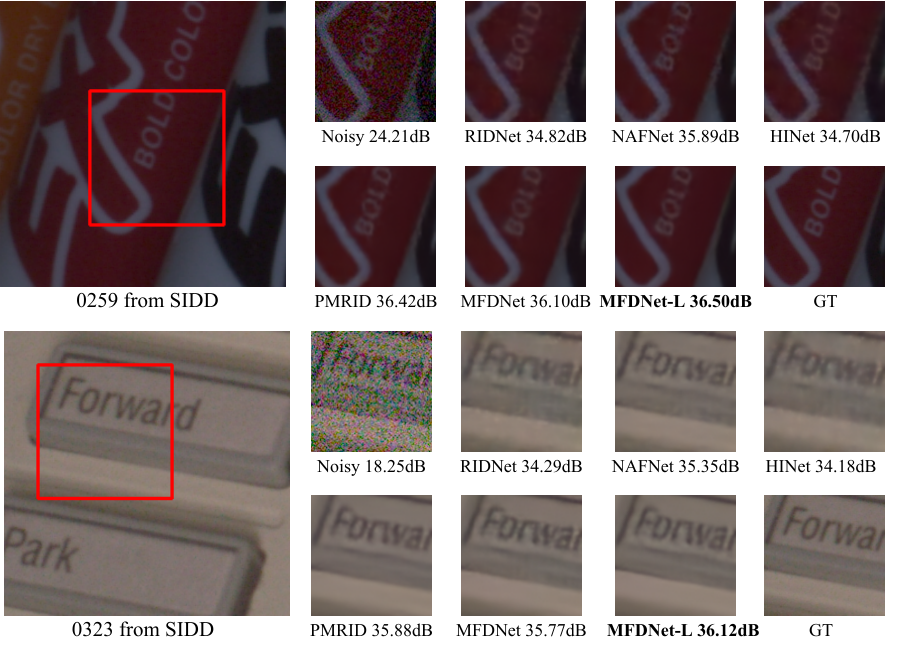}
    \caption{Qualitative comparisons on SIDD.}
    \label{fig:my_label}
\end{figure}

\begin{table}[]\scriptsize
\caption{Quantitative performance of different methods.}\vspace{0.2cm}
\label{denoising and runtime performance}
\centering
\setlength{\tabcolsep}{0.5mm}{
\begin{tabular}{|ccccccccc|}
\hline
\multicolumn{1}{|c|}{\multirow{2}{*}{\makecell[c]{\vspace{-0.05cm}\\Model}}} & \multicolumn{1}{c|}{\multirow{2}{*}{\makecell[c]{\vspace{-0.1cm}\\ MACs \\ /G}}} & \multicolumn{1}{c|}{\multirow{2}{*}{\makecell[c]{\vspace{-0.1cm}\\ Memory \\ /M}}} & \multicolumn{2}{c|}{Latency/ms}  & \multicolumn{2}{c|}{SIDD}                               & \multicolumn{2}{c|}{DND}           \\ \cline{4-9} 
\multicolumn{1}{|c|}{}                       & \multicolumn{1}{c|}{}                                                                   & \multicolumn{1}{c|}{}                                                                     & \multicolumn{1}{c|}{iPhone 11} & \multicolumn{1}{c|}{\begin{tabular}[c]{@{}c@{}}iPhone \\ 14 Pro \end{tabular}} & \multicolumn{1}{c|}{PSNR}  & \multicolumn{1}{c|}{SSIM}  & \multicolumn{1}{c|}{PSNR}  & SSIM  \\ \hline
\multicolumn{1}{|c|}{DnCNN-S \cite{zhang2017beyond}}                & \multicolumn{1}{c|}{$2.77$}                                                               & \multicolumn{1}{c|}{$583$}                                                                  & \multicolumn{1}{c|}{$11.7$}      & \multicolumn{1}{c|}{$5.7$}                                                          & \multicolumn{1}{c|}{$33.84$} & \multicolumn{1}{c|}{$0.877$} & \multicolumn{1}{c|}{$36.41$} & $0.910$ \\ \hline
\multicolumn{1}{|c|}{NAFNet \cite{chen2022simple}}                 & \multicolumn{1}{c|}{$3.81$}                                                               & \multicolumn{1}{c|}{$2974$}                                                                 & \multicolumn{1}{c|}{$1384.5^*$}    & \multicolumn{1}{c|}{$25.9$}                                                         & \multicolumn{1}{c|}{$\bf{38.66}$} & \multicolumn{1}{c|}{$\bf{0.951}$} & \multicolumn{1}{c|}{$\bf{38.74}$} & $\bf{0.945}$ \\ \hline
\multicolumn{1}{|c|}{MFDNet-S}               & \multicolumn{1}{c|}{$2.34$}                                                               & \multicolumn{1}{c|}{$142$}                                                                  & \multicolumn{1}{c|}{$\bf{8.3}$}       & \multicolumn{1}{c|}{$\bf{4.0}$}                                                          & \multicolumn{1}{c|}{$36.93$} & \multicolumn{1}{c|}{$0.932$} & \multicolumn{1}{c|}{$38.09$} & $0.939$ \\ \hline
                                             &                                                                                         &                                                                                           &                                &                                                                                   &                            &                            &                            &       \\ \hline
\multicolumn{1}{|c|}{DnCNN \cite{zhang2017beyond}}                  & \multicolumn{1}{c|}{$12.88$}                                                              & \multicolumn{1}{c|}{$2720$}                                                                 & \multicolumn{1}{c|}{$23.4$}      & \multicolumn{1}{c|}{$12.5$}                                                         & \multicolumn{1}{c|}{$36.24$} & \multicolumn{1}{c|}{$0.921$} & \multicolumn{1}{c|}{$38.13$} & $0.936$ \\ \hline
\multicolumn{1}{|c|}{HINet \cite{chen2021hinet}}                  & \multicolumn{1}{c|}{$11.68$}                                                              & \multicolumn{1}{c|}{$2178$}                                                                 & \multicolumn{1}{c|}{$84.7^*$}      & \multicolumn{1}{c|}{$23.3$}                                                         & \multicolumn{1}{c|}{$37.83$} & \multicolumn{1}{c|}{$0.943$} & \multicolumn{1}{c|}{$36.82$} & $0.931$ \\ \hline
\multicolumn{1}{|c|}{MFDNet}                 & \multicolumn{1}{c|}{$11.46$}                                                              & \multicolumn{1}{c|}{$384$}                                                                  & \multicolumn{1}{c|}{$\bf{14.2}$}      & \multicolumn{1}{c|}{$\bf{6.2}$}                                                          & \multicolumn{1}{c|}{$\bf{38.90}$} & \multicolumn{1}{c|}{$\bf{0.952}$} & \multicolumn{1}{c|}{$\bf{39.06}$} & $\bf{0.947}$ \\ \hline
                                             &                                                                                         &                                                                                           &                                &                                                                                   &                            &                            &                            &       \\ \hline
\multicolumn{1}{|c|}{RIDNet \cite{zhuo2019ridnet}}                 & \multicolumn{1}{c|}{$20.39$}                                                              & \multicolumn{1}{c|}{$4224$}                                                                 & \multicolumn{1}{c|}{$75.9$}      & \multicolumn{1}{c|}{$30.0$}                                                         & \multicolumn{1}{c|}{$38.01$} & \multicolumn{1}{c|}{$0.945$} & \multicolumn{1}{c|}{$38.88$} & $0.947$ \\ \hline
\multicolumn{1}{|c|}{PMRID \cite{wang2020practical}}                  & \multicolumn{1}{c|}{$15.07$}                                                              & \multicolumn{1}{c|}{$4448$}                                                                 & \multicolumn{1}{c|}{$80.8$}      & \multicolumn{1}{c|}{$40.1$}                                                         & \multicolumn{1}{c|}{$38.96$} & \multicolumn{1}{c|}{$0.953$} & \multicolumn{1}{c|}{$38.82$} & $0.948$ \\ \hline
\multicolumn{1}{|c|}{MFDNet-L}               & \multicolumn{1}{c|}{$21.81$}                                                              & \multicolumn{1}{c|}{$684$}                                                                  & \multicolumn{1}{c|}{$\bf{24.1}$}      & \multicolumn{1}{c|}{$\bf{8.9}$}                                                          & \multicolumn{1}{c|}{$\bf{39.01}$} & \multicolumn{1}{c|}{$\bf{0.954}$} & \multicolumn{1}{c|}{$\bf{39.15}$} & $\bf{0.948}$ \\ \hline
\end{tabular}}
\end{table}

\section{Conclusion}\vspace{-0.3cm}

In this paper, we identify the network architectures and operations that can run on NPUs with low latency and excellent denoising performance through extensive analysis and experiments. Based on that, we build a mobile-friendly denoising network from scratch. Experiments show advances of our method in terms of both denoising and runtime performance. We hope this work will promote the application of CNN-based denoising models on mobile devices.

\end{document}